\documentclass[final]{york-thesis}

\usepackage{url}
\usepackage{array}
\usepackage{amsmath}
\usepackage{amssymb}
\usepackage{graphicx}
\usepackage{multirow}
\usepackage{CJKutf8}

\usepackage{fancyhdr}
\renewcommand\headheight{24pt}

\title{A Study of the Tasks and Models in \ \ \ \ \ \ \ \ \ \ \ \ \ \ \ \ \ Machine Reading Comprehension}
\author{Chao Wang}
\degreename{Doctor}
\department{Electrical Engineering and Computer Science}
\date{October 2019}
\setboolean{masters}{false}

\begin{document}
\abstractfile{abstract}
\makefrontmatter
\chapter{Introduction}
Machine Reading Comprehension (MRC) requires a machine to read a context and answer a set of relevant questions based on its comprehension of the context. As a challenging area in natural language processing (NLP), MRC has attracted tremendous interest from the artificial intelligence community. In recent years, many MRC tasks have been established to facilitate the explorations and innovations in this area. These tasks vary a lot in both dataset collection and performance evaluation, but in this report, they are divided roughly into two categories according to the complexity of the required reasoning process:
\begin{itemize}
\item Simple-reasoning MRC tasks, where each context is a single passage, such as a single fictional story or newspaper article, so that the required reasoning process is relatively simple.
\item Complex-reasoning MRC tasks, where each context consists of multiple passages, such as multiple book chapters or web documents, so that the required reasoning process is relatively complex.
\end{itemize}
Due to the rapid advance and great success of deep learning, neural networks have been the dominant techniques for developing MRC models to address MRC tasks. The applications of neural networks make it necessary for MRC models to generate word representations for each given context-question pair, which is usually an iterative process across multiple abstraction levels. In this way, the core components of MRC models are their attention mechanisms \cite{bahdanau2014neural}, which are aimed at integrating the representations of semantically associated words in one abstraction level to generate the representations in the next abstraction level. In this report, the attention mechanisms applied to MRC models are divided roughly into two categories: mutual-matching attention, which is aimed at fusing the question representations into the context representations to generate the question-aware context representations, and self-matching attention, which is aimed at fusing the question-aware context representations into themselves to generate the final context representations. To capture the interactions among words more effectively, attention mechanisms can be performed repeatedly, which is known as multi-round attention. Besides attention mechanisms, many performance-boosting approaches, such as linguistic embeddings, multi-round reasoning, reinforcement learning, and data augmentation, have also been applied to MRC models. \\
To bridge the gap between MRC models and human beings, it is helpful to incorporate external knowledge into neural networks. Two styles of external knowledge are involved in this report, namely text-style knowledge contained in external corpora (e.g. Wikipedia) and graph-style knowledge contained in external knowledge bases (e.g. Freebase \cite{bollacker2008freebase}). To incorporate text-style external knowledge into neural networks, an effective way is to conduct transfer learning by pre-training a source model through a foundation-level NLP task based on external corpora. To incorporate graph-style external knowledge into neural networks, an effective way is to encode external knowledge bases in vector space such that the encoding outputs can be used to enhance word representations. \\
This report focuses on the major characteristics of the existing MRC tasks and the major techniques applied to the existing MRC models. This chapter is a general introduction. Chapter 2 introduces some representative simple-reasoning and complex-reasoning MRC tasks. Chapter 3 introduces some representative architecture designs, attention mechanisms, and performance-boosting approaches for MRC models. Chapter 4 introduces some representative applications of transfer learning in MRC models. Chapter 5 introduces some representative applications of knowledge base encoding in MRC models. Chapter 6 is a conclusion that summarizes the report and proposes some open problems.

\chapter{An Overview of MRC Tasks}
There are many categories of NLP tasks, such as sentiment analysis, part-of-speech tagging, named-entity recognition, to name a few. Compared with other NLP tasks, MRC tasks are more demanding in techniques, more concerned with serving real-world users, and also more relevant to the ultimate target of artificial intelligence. This chapter introduces some representative MRC tasks, which cover both simple-reasoning and complex-reasoning MRC tasks.

\section{Simple-reasoning MRC Tasks}
This section introduces the dataset collection and performance evaluation of some representative simple-reasoning MRC tasks.

\subsection{MCTest}
\begin{center}
\fbox{
\parbox{0.95\textwidth}{
\textbf{Context}: James the Turtle was always getting in trouble. Sometimes he'd reach into the freezer and empty out all the food. Other times he'd sled on the deck and get a splinter. His aunt Jane tried as hard as she could to keep him out of trouble, but he was sneaky and got into lots of trouble behind her back... \\
\textbf{Question}: What is the name of the trouble making turtle? \\
\textbf{Candidate Answers}: 1) Fries; 2) Pudding; 3) James; 4) Jane
}
}
\end{center}
MCTest \cite{richardson2013mctest} is a choice-style MRC task, where each context is a single passage attached with several questions, each question has several candidate answers, and the target is to determine which candidate answer is correct. The dataset of MCTest contains $660$ contexts, $2640$ questions, and $10560$ candidate answers, which were all produced by crowd-sourcing workers. First, the crowd-sourcing workers were asked to write fictional stories that can be easily understood by school children to form the contexts. As a result, the information necessary to answer each question only exists in the corresponding context, and the effect of background knowledge in determining the correct answer is very limited. Then, the crowd-sourcing workers were asked to write four questions for each context and four candidate answers for each question with each candidate answer containing words from the context. As a result, even the incorrect answers look reasonable such that the questions are not easy to tackle. Finally, the crowd-sourcing workers were asked to make sure that at least two questions for each context require multiple sentences from the context to determine the correct answer. As a result, MRC models cannot benefit too much from simple word matching. MCTest is the first MRC task that directly focuses on the top-level goals of MRC, such as causal reasoning and world understanding. By adopting the choice style, MCTest is able to use an objective metric to evaluate the progress towards these goals. However, MCTest is too small in dataset scale to support the development of data-intensive MRC models.

\subsection{WikiQA}
\begin{center}
\fbox{
\parbox{0.95\textwidth}{
\textbf{Question}: Who wrote Second Corinthians? \\
\textbf{Candidate Answer Sentences} (the sentences in the summary paragraph of the Wikipedia page ``Second Epistle to the Corinthians''): \\
1) The Second Epistle to the Corinthians, commonly referred to as Second Corinthians or in writing 2 Corinthians, is a Pauline epistle of the New Testament of the Christian Bible. \\
2) The epistle is attributed to Paul the Apostle and a co-author named Timothy, and is addressed to the church in Corinth and Christians in the surrounding province of Achaea, in modern-day Greece.
}
}
\end{center}
WikiQA \cite{yang2015wikiqa} is another choice-style MRC task, or more specifically, an answer sentence selection task, where each question is attached with several candidate answer sentences, which can be seen as the context, and the target is to determine which candidate answer sentence is correct. The dataset of WikiQA contains $3047$ questions, which were collected from the Bing query logs in a heuristic way such that each question was both frequently issued by users and associated with a Wikipedia page. For each question, the sentences in the summary paragraph of its associated Wikipedia page were used as its candidate answer sentences, which formed $29258$ candidate answer sentences. As a result, the questions and the candidate answer sentences are very similar in distribution to those in the real world. Crowd-sourcing workers were recruited to label whether each candidate answer sentence is correct or not. To facilitate the research on answer triggering, the questions without any candidate answer sentence labeled as correct, which took almost two-thirds of all questions, were not filtered out but kept in the dataset. As a result, MRC models need to know if a question is answerable or not, which is also the case in the real world. WikiQA uses a group of well-defined metrics for performance evaluation. On the questions having correct answer sentences, WikiQA uses Mean Average Precision (MAP) and Mean Reciprocal Rank (MRR) to evaluate the answer sentence selection performance. On all the questions, WikiQA uses question-level precision, question-level recall, and question-level F1 score to evaluate the answer triggering performance. In terms of dataset scale, like MCTest, WikiQA is also too small to support the development of data-intensive MRC models.

\subsection{CNN/DailyMail}
\begin{center}
\fbox{
\parbox{0.95\textwidth}{
\textbf{Context} (anonymised): The ent381 producer allegedly struck by ent212 will not press charges against the ``ent153'' host, his lawyer said Friday. ent212, who hosted one of the most-watched television shows in the world, was dropped by the ent381 Wednesday after an internal investigation by the ent180 broadcaster found he had subjected producer ent193 ``to an unprovoked physical and verbal attack.''... \\
\textbf{Question} (anonymised): producer $\rule{1cm}{0.15mm}$ will not press charges against ent212, his lawyer says.
}
}
\end{center}
CNN/DailyMail \cite{hermann2015teaching} is a cloze-style MRC task, where each context is a single passage attached with several extra sentences, each of which contains a placeholder and thus can be seen as a question, and the target is to fill in each placeholder with an entity contained in the context. The dataset of CNN/DailyMail contains $93000$ contexts collected from the online newspaper articles of CNN and $220000$ from Daily Mail. Since CNN and Daily Mail supplemented each of their articles with several bullet points, these bullet points were used to generate the questions by replacing one entity in each bullet point with a placeholder. To eliminate the effect of background knowledge and focus on the core challenges in MRC, CNN/DailyMail anonymised all mentions of each entity with a randomly numbered abstract marker. While CNN/DailyMail is the first MRC task with a large-scale dataset, its dataset is noisy due to the errors caused by its automatic data generation procedure, and the complexity of its required reasoning process is low \cite{chen2016thorough}.

\subsection{Children's Book Test}
\begin{center}
\fbox{
\parbox{0.95\textwidth}{
\textbf{Context}: \\
1) Mr. Cropper was opposed to our hiring you. \\
2) Not, of course, that he had any personal objection to you, but he is set against female teachers, and when a Cropper is set there is nothing on earth can change him. \\
... \\
20) Esther felt relieved. \\
\textbf{Question}: She thought that Mr. $\rule{1cm}{0.15mm}$ had exaggerated matters a little. \\
\textbf{Candidate Answers}: 1) Baxter; 2) Cropper; ... 10) spite
}
}
\end{center}
Children's Book Test (CBT) \cite{hill2015goldilocks} is another cloze-style MRC task, where a machine is given an ordered list of sentences with the last sentence containing a missing word symbol, and required to identify the missing word from a set of candidate words. Therefore the preceding sentences can be seen as the context, the last sentence as the question, and the missing word as the answer. The dataset of CBT contains $687343$ context-question pairs, which were collected from $108$ children's books that are freely available in Project Gutenberg. Specifically, each context-question pair was generated by enumerating $21$ consecutive sentences in a book chapter, and attached with $10$ candidate words appearing in the context-question pair. There are four types of missing words, namely entities, nouns, verbs, and prepositions. For a given missing word, only the words in the corresponding context-question pair with the same type will be considered as qualified candidate words, and a heuristic algorithm was used to deal with the situation where there are insufficient words of a given type. Although there are many similarities between CBT and CNN/DailyMail, their differences are obvious. On the one hand, in CNN/DailyMail the answers were limited to entities, but in CBT there were more types. On the other hand, unlike CNN/DailyMail, CBT did not anonymise entities and therefore incentivise the application of background knowledge.

\subsection{SQuAD}
\begin{center}
\fbox{
\parbox{0.95\textwidth}{
\textbf{Context}: In meteorology, precipitation is any product of the condensation of atmospheric water vapour that falls under gravity. The main forms of precipitation include drizzle, rain, sleet, snow, graupel and hail... \\
\textbf{Question}: What causes precipitation to fall?
}
}
\end{center}
SQuAD \cite{rajpurkar2016squad} is a span-style MRC task, where each context is a single passage attached with several questions, and the target is to predict an answer span (i.e. the answer start position and the answer end position) from the context for each question. The dataset of SQuAD contains $100000$ context-question pairs labeled with answer spans, where the contexts were collected from Wikipedia, while the questions and the answers were produced by crowd-sourcing workers. To examine the robustness to noise of MRC models, two noise-injected adversarial sets of SQuAD, namely AddSent and AddOneSent \cite{jia2017adversarial}, have also been released. The contexts in the adversarial sets contain machine-generated misleading sentences, which are aimed at distracting MRC models. Specifically, each context in AddSent contains several sentences that are similar to the question but not contradictory to the answer, while each context in AddOneSent contains a human-approved random sentence that is usually unrelated to the context. It has been revealed that when evaluated on the above noisy adversarial sets, many existing MRC models drop significantly in the performance. To facilitate the research on answer triggering, $53775$ unanswerable questions with plausible answers, which were also produced by crowd-sourcing workers, were combined into the original dataset so that SQuAD 2.0 \cite{rajpurkar2018know} was established. In SQuAD 2.0, a machine must not only answer questions when possible, but also determine when no answer is supported by the context so as to abstain from answering. It has been revealed that when transplanted to SQuAD 2.0, many existing MRC models that perform well on SQuAD drop significantly in the performance. For performance evaluation, both SQuAD and SQuAD 2.0 adopt Exact Match (EM) and F1 Score as the evaluation metrics.

\subsection{NewsQA}
\begin{center}
\fbox{
\parbox{0.95\textwidth}{
\textbf{Context}: Hyrule (Contoso) -- Tingle, Tingle! Kooloo-Limpah! ...These are the magic words that Tingle created himself. Don't steal them! \\
\textbf{Question}: What should you not do with Tingle's magic words?
}
}
\end{center}
NewsQA \cite{trischler2016newsqa} is another span-style MRC task. Like in SQuAD, the dataset of NewsQA contains $119633$ context-question pairs labeled with answer spans, where the contexts were collected from $12744$ online newspaper articles of CNN, while the questions and the answers were produced by crowd-sourcing workers. Besides, like in SQuAD 2.0, a significant amount of the questions in NewsQA are unanswerable and thus labeled with a null span. However, NewsQA is different from SQuAD in dataset collection. On the one hand, the crowd-sourcing workers responsible for writing questions are separated from those for labeling answer spans, which raises the difficulty of questions. On the other hand, the question-writing workers were given the bullet points instead of the content of each context, which encourages their curiosity, prevents questions that are reformulations of context sentences, and also increases the amount of unanswerable questions. Due to the above characteristics in dataset collection, NewsQA requires MRC models to possess reasoning abilities that are beyond word matching and context matching.

\section{Complex-reasoning MRC Tasks}
This section introduces the dataset collection and performance evaluation of some representative complex-reasoning MRC tasks.

\subsection{MS MARCO}
\begin{center}
\fbox{
\parbox{0.95\textwidth}{
\textbf{Context Passage 1}: Results-Based Accountability (also known as RBA) is a disciplined way of thinking and taking action that communities can use to improve the lives of children, youth, families, adults and the community as a whole... \\
... \\
\textbf{Context Passage 10}: Get To Know Us. RBA is a digital and technology consultancy with roots in strategy, design and technology... \\
\textbf{Question}: What is RBA?
}
}
\end{center}
MS MARCO \cite{bajaj2016ms} is a generation-style MRC task, where each question is attached with a multi-passage context with the passages collected from different sources, and the target is to generate an abstractive summary for both the question and the context as the answer. The dataset of MS MARCO contains $100000$ context-question pairs labeled with human-generated answers, where the questions were real-world user queries issued to Bing, the context passages for each question were extracted from the corresponding top-$10$ web pages retrieved by Bing, and the answers were produced by crowd-sourcing workers. Based on various answer types, MS MARCO divided the questions into five categories, namely description, numeric, entity, person, and location. To facilitate the research on answer triggering, MS MARCO retained all unanswerable questions so that MRC models are required to determine whether to abstain from answering. Besides, inspired by the reasoning ability of human beings, MS MARCO labeled each context-question pair with the supporting passages that are useful for generating the answer so that MRC models are also required to predict them. For performance evaluation, MS MARCO adopts ROUGE-L and BLEU-1 as the evaluation metrics for answer generation, and adopts MAP and MRR for supporting passage prediction.

\subsection{DuReader}
\begin{center}
\fbox{
\parbox{0.95\textwidth}{
\textbf{Context Passage 1}: \begin{CJK*}{UTF8}{gbsn}为什么要拔智齿? 智齿好好的医生为什么要建议我拔掉? 主要还是因为智齿很难清洁...\end{CJK*} \\
... \\
\textbf{Context Passage 5}: \begin{CJK*}{UTF8}{gbsn}根据我多年的临床经验来说, 智齿不一定非得拔掉。 智齿阻生分好多种...\end{CJK*} \\
\textbf{Question}: \begin{CJK*}{UTF8}{gbsn}智慧牙一定要拔吗?\end{CJK*}
}
}
\end{center}
DuReader \cite{he2017dureader} is another generation-style MRC task. For dataset collection, DuReader first used a pre-trained classifier to select questions from the most frequent user queries issued to Baidu; then queried Baidu Search or Baidu Zhidao with each obtained question so as to extract context passages from the top-$5$ retrieved web pages; next asked crowd-sourcing workers to annotate each obtained question as either Fact or Opinion, and one of Entity, Description, and YesNo; and finally asked crowd-sourcing workers to produce answers in their own words. As a result, the dataset of DuReader contains $200000$ questions, $1000000$ context passages, and more than $420000$ answers, which are all in Chinese. Unlike MS MARCO, where each context passage is a single paragraph, DuReader provides unabridged articles as context passages, therefore the context passages in DuReader are $5$ times longer than those in MS MARCO. For performance evaluation, DuReader adopts ROUGE-L and BLEU-4 as the evaluation metrics.

\subsection{NarrativeQA}
\begin{center}
\fbox{
\parbox{0.95\textwidth}{
\textbf{Context} (a whole movie script or book): ...She continues digging in her purse while Frank leansover the buggy and makes funny faces at the baby, OSCAR, a very cute nine-month old boy... \\
\textbf{Context Summary}: ...Peter's former girlfriend Dana Barrett has had a son, Oscar... \\
\textbf{Question}: How is Oscar related to Dana?
}
}
\end{center}
NarrativeQA \cite{kovcisky2018narrativeqa} is another generation-style MRC task, where the contexts are books and movie scripts. For dataset collection, NarrativeQA first screened the books from Project Gutenberg and the movie scripts from Internet such that only the books and movie scripts that have corresponding summaries in Wikipedia were kept; then asked crowd-sourcing workers to write question-answer pairs based on the summaries; and finally asked a different crowd-sourcing worker to write a second answer for each question if the worker believe the question to be answerable. As a result, the dataset of NarrativeQA contains $1567$ contexts and $46765$ question-answer pairs. Since the crowd-sourcing workers only saw the summaries instead of the books or movie scripts, they were unlikely to derive question-answer pairs from localized snippets so that MRC models need to understand the underlying narrative rather than just rely on superficial information. Another challenge of NarrativeQA lies in the fact that the contexts are very long, which makes it computationally infeasible for most existing MRC models. For performance evaluation, NarrativeQA adopts BLEU-1, BLEU-4, Meteor, and ROUGE-L as the evaluation metrics for answer generation, and MRR as a ranking metric.

\subsection{SearchQA}
\begin{center}
\fbox{
\parbox{0.95\textwidth}{
\textbf{Context Passage 1}: The Klingons are a fictional extraterrestrial humanoid warrior species in the science fiction ... \\
... \\
\textbf{Context Passage 94}: I have a strong prediction of what the number one spot will be... \\
\textbf{Question}: Guinness says that by number of users this language, devised by James Doohan, is the most spoken fictional language.
}
}
\end{center}
SearchQA \cite{dunn2017searchqa} is a span-style MRC task, where each question is attached with a multi-passage context with the passages collected from different sources, and the target is to predict an answer span (i.e. the answer start position and the answer end position) from the context. For dataset collection, SearchQA first crawled the entire set of question-answer pairs from J! Archive; then queried Google with each obtained question so as to extract context passages from the retrieved web pages; and finally cleaned the resulting question-answer-context tuples in a heuristic way such that each answer is no longer than three words, contained in the corresponding context, and unlikely to be found trivially through word matching. As a result, the dataset of SearchQA contains $140461$ question-answer-context tuples with each context consisting of $49.6$ passages on average. For performance evaluation, SearchQA adopts top-$1$ accuracy and top-$5$ accuracy as the evaluation metrics for multi-word answers, and adopts F1 score for single-word answers.

\subsection{TriviaQA}
\begin{center}
\fbox{
\parbox{0.95\textwidth}{
\textbf{Context Passage 1}: The Nobel Prize in Literature 1930 was awarded to Sinclair Lewis ``for his vigorous and graphic art of description and his ability to create, with wit and humour, new types of characters.'' \\
... \\
\textbf{Context Passage 7}: SINCLAIR LEWIS (1885-1951) was an American novelist, short-story writer, and playwright... \\
\textbf{Question}: Which American-born Sinclair won the Nobel Prize for Literature in 1930?
}
}
\end{center}
TriviaQA \cite{joshi2017triviaqa} is another span-style MRC task. For dataset collection, TriviaQA first crawled the question-answer pairs with the question length no less than four words from several trivia websites; then queried Bing with each obtained question so as to retrieve the top-$50$ relevant web pages; and finally extract context passages from the top $10$ of the resulting web pages that are not from the trivia websites. As a result, the dataset of TriviaQA contains $95000$ question-answer-context tuples with each context consisting of $6$ passages on average. TriviaQA is challenging in that a large proportion of its questions have highly compositional semantics, exhibit substantial syntactic and lexical variability when compared with the corresponding contexts, and require reasoning over multiple sentences. Besides, it is also worth noting that TriviaQA provided answer strings instead of answer spans as labels, which makes it necessary to apply distant supervision since a presence of a answer string is not guaranteed to imply the answer. For performance evaluation, TriviaQA adopts EM and F1 Score as the evaluation metrics.

\subsection{HotpotQA}
\begin{center}
\fbox{
\parbox{0.95\textwidth}{
\textbf{Context Passage 1}: The 1995–96 season was Manchester United's fourth season in the Premier League, and their 21st consecutive season in the top division of English football... \\
... \\
\textbf{Context Passage 10}: Sir Alexander Chapman Ferguson, CBE (born 31 December 1941) is a Scottish former football manager and player who managed Manchester United from 1986 to 2013... \\
\textbf{Question}: The football manager who recruited David Beckham managed Manchester United during what timeframe?
}
}
\end{center}
HotpotQA \cite{yang2018hotpotqa} is another span-style MRC task, where the contexts are summary paragraphs of Wikipedia and the answers could be yes or no besides spans. For dataset collection, HotpotQA first sampled paragraph pairs from the summary paragraphs of Wikipedia; then asked crowd-sourcing workers to come up with a question for each paragraph pair such that answering the question requires reasoning between the paragraph pair; next asked the crowd-sourcing workers to label not only the answer but also the supporting facts (i.e. the sentences that are necessary to arrive at the answer) on the paragraph pair for each question; and finally mixed each paragraph pair with $8$ distracting paragraphs to form the context passages. As a result, the dataset of HotpotQA contains $112779$ question-answer-context tuples with each context consisting of $10$ passages. The questions in HotpotQA fall into two categories, namely bridge questions and comparison questions, which are different in the way the corresponding paragraph pairs were sampled. For bridge questions, the summary paragraphs of Wikipedia were organized into a directed graph with each edge representing a hyperlink from the source paragraph to the destination paragraph, and the paragraph pairs were sampled from the edges of the graph with the destination paragraphs constrained to certain scope. For comparison questions, the summary paragraphs of Wikipedia were organized into a set of lists according to their title entities, and each paragraph pair was sampled from the same list. By the way, to increase the diversity of questions, a subset of the comparison questions were designed as yes-or-no questions. The challenge of HotpotQA is two-fold, on the one hand, answering a question requires reasoning over multiple sentences distributed in different paragraphs, which is known as multi-hop reasoning; on the other hand, since both the answers and the supporting facts were provided as strong supervision, MRC models are expected to be capable of reasoning in an explainable manner. For performance evaluation, HotpotQA adopts EM and F1 Score as the evaluation metrics for both answer prediction and supporting facts prediction, and combines the evaluation metrics for the two tasks to get Joint EM and Joint F1 Score for an overall evaluation.

\chapter{An Overview of MRC Models}
Developing a neural-network-based MRC model is not a trivial job. First of all, the MRC model needs to adopt reasonable architecture designs. Then, the major difficulty lies in properly using attention mechanisms to integrate the representations of semantically associated words. Last but not least, the MRC model also needs to apply appropriate performance-boosting approaches. This chapter introduces these primary aspects of MRC model development.

\section{Architecture Designs for MRC Models}
This section introduces some representative architecture designs for MRC models, which cover both simple-reasoning and complex-reasoning MRC models.

\subsection{Simple-reasoning MRC Models}
\begin{figure*}
\centering
\includegraphics[width=0.5\linewidth]{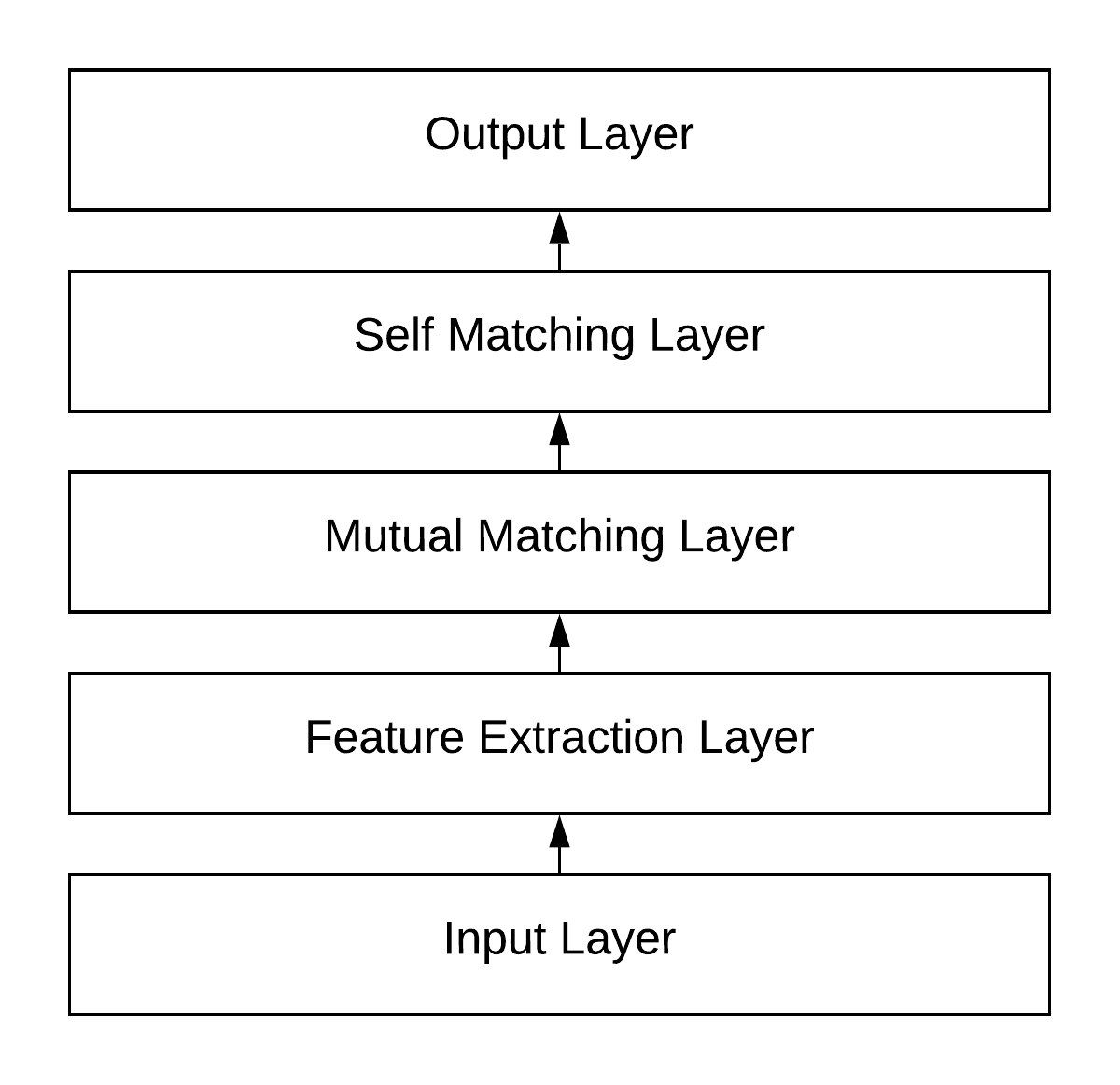}
\end{figure*}
As shown in the figure, a conceptual architecture of simple-reasoning MRC models consists of five layers, which are separately: \\
\textbf{Input Layer.} This layer maps the words in each given context-question pair to the lexical embeddings, which are usually assigned the corresponding word vectors provided by a pre-trained Word2Vec \cite{mikolov2013distributed} or GloVe \cite{pennington2014glove}. For out-of-vocabulary (OOV) words with low frequency, their lexical embeddings are usually assigned the same random or zero vector. To better deal with OOV words, it is a common practice to append a character-level embedding to the lexical embedding of each word, which is usually obtained by processing the corresponding character vectors with a Convolutional Neural Network (CNN) or Recurrent Neural Network (RNN). Besides, some MRC models, such as DrQA \cite{chen2017reading}, R.M-Reader \cite{hu2017reinforced}, and SAN \cite{liu2017stochastic}, also append binary features to the lexical embedding of each context word indicating whether this word and its variants appear in the question. \\
\textbf{Feature Extraction Layer.} This layer maps the lexical embeddings to the contextual embeddings, which is aimed at extracting contextual features from the context and the question separately. A common practice for implementing this layer is to process the lexical embeddings of the context and those of the question separately with a shared bidirectional LSTM (BiLSTM) \cite{hochreiter1997long} or bidirectional GRU (BiGRU) \cite{cho2014properties}, and thereby concatenate the outputs from both directions. To improve the training and inference efficiency, some MRC models, such as QANet \cite{yu2018qanet}, replace all recurrent structures with feed-forward structures. \\
\textbf{Mutual-matching Layer.} This layer matches the contextual embeddings of the context with those of the question to generate the coarse memories of the context, which is also known as question-aware context representations. Mutual-matching attention is the substantial component of this layer, which fuses the contextual embeddings of the question into those of the context according to their mutual similarities. The fusion results are usually passed through a BiLSTM or BiGRU to form the coarse memories of the context. Some MRC models, such as SLQA+ \cite{wangwei2018multi}, also generate the coarse memories of the question in a symmetrical manner, which can be seen as context-aware question representations. \\
\textbf{Self-matching Layer.} This layer matches the coarse memories of the context with themselves to generate the refined memories of the context, which is the self-aware context representations. Self-matching attention is the substantial component of this layer, which fuses the coarse memories of the context into themselves according to their self similarities. The fusion results are usually passed through a BiLSTM or BiGRU to form the refined memories of the context. \\
\textbf{Output Layer.} This layer predicts certain distributions, such as the distributions over the candidate answers, the context words, and so on, as outputs, the format of which depends on the specific MRC task to address. \\
The above architecture is just a general framework for constructing simple-reasoning MRC models, which is more or less different from those applied in practice. For example, Match-LSTM \cite{wang2016machine}, RaSoR \cite{lee2016learning}, DCN \cite{xiong2016dynamic}, BiDAF \cite{seo2016bidirectional}, MEMEN \cite{pan2017memen}, DrQA \cite{chen2017reading}, FastQA \cite{weissenborn2017making}, and QANet \cite{yu2018qanet} have no self-matching layer, while DrQA and FastQA even have no mutual-matching layer.

\subsection{Complex-reasoning MRC Models}
Complex-reasoning MRC models usually borrow a lot of architecture designs from simple-reasoning MRC models, and also develop new techniques to provide new functions, such as screening out distracting passages, generating textual answers, and reasoning across passages, to name a few. \\
In most complex-reasoning MRC tasks, only a few passages of each context are relevant to the corresponding question, while the rest are useless and thus could distract MRC models, therefore it is necessary to screen out the distracting passages so as to focus on the relevant ones. To this end, a confidence approach \cite{clark2017simple} first uses a passage ranking model to select the most relevant passages from each context, then samples a subset from the selected passages for training and the full set for inference, next uses a simple-reasoning MRC model to generate an unnormalized answer span distribution over each sampled passage, and finally uses a shared normalization to generate a normalized answer span distribution over all the sampled passages. A deep cascade approach \cite{yan2019deep} first uses a document ranking model and a paragraph ranking model to select the most relevant passages (i.e. the most relevant paragraphs in the most relevant documents), then adapts a simple-reasoning MRC model to the question and the selected passages by processing different passages in parallel, and finally performs document extraction and paragraph extraction as two auxiliary tasks of answer extraction. The above approaches screen out a part of the distracting passages in advance through a ranking process, which is necessary when each context consists of many passages. But in cases where the number of passages per context is limited, the ranking process can be skipped and the distracting passages can be screened out by simply performing passage extraction together with answer extraction in a multi-task manner \cite{tan2017s}. \\
In some complex-reasoning MRC tasks, golden answers are not necessarily exact spans in contexts but human-generated abstractive summaries of context-question pairs. Although such abstractive answers can be approximated by training MRC models to predict the most similar answer spans \cite{wang2018multi, hu2018attention}, it is more interesting to train MRC models to directly generate textual answers. To this end, an extraction-then-synthesis approach \cite{tan2017s} first trains an MRC model to extract an answer span based on each given context-question pair, and then trains a text generation model to synthesize a textual answer based on the context-question pair and the extracted answer span. A multi-style abstractive summarization approach \cite{nishida2019multi} solved this problem by proposing an end-to-end MRC model, which applies a Transformer \cite{vaswani2017attention} based pointer-generator mechanism \cite{see2017get} such that each answer word is either drawn from the vocabulary or copied from the corresponding context-question pair, and also sets special tokens at the answer start position to represent answer styles, such as well-formed sentences and concise phrases. \\
In some complex-reasoning MRC tasks, the answer to each question may appear in multiple passages of the corresponding context, since the passages of the context are independent from each other. Besides, the appearance of the answer in a passage does not imply that the passage alone is sufficient for the question, since locating answer may require information from other passages. Therefore, reasoning across passages to integrate supporting information can usually improve the performance of complex-reasoning MRC models. An answer re-ranking approach \cite{wang2017evidence} first collects the most probable answer candidates from a baseline MRC model, and then uses two proposed re-rankers, namely strength-based re-ranker and coverage-based re-ranker, to re-score the answer candidates by aggregating evidences from the corresponding passages for each answer candidate. A cross-passage answer verification approach \cite{wang2018multi} performs answer content modeling and answer verification as two auxiliary tasks of answer extraction, where answer content modeling is estimating whether each word should be included in the answer, and answer verification is verifying the answer from each passage with those from the other passages to estimate its correctness. An answering-while-summarizing approach \cite{nishida2019answering} uses a proposed query-focused extractor to perform evidence extraction as an auxiliary task of answer extraction, which sequentially extracts evidence sentences from the context sentences with an RNN attending to the question sentence such that the sentence-level relationship is taken into account.

\section{Attention Mechanisms for MRC Models}
This section introduces some representative attention mechanisms for MRC models.

\subsection{Mutual-matching Attention}
As the substantial component of the mutual-matching layer, mutual-matching attention is aimed at fusing the contextual embeddings of the question, $U_Q \in \mathbb{R}^{d \times m}$, into those of the context, $U_C \in \mathbb{R}^{d \times n}$, where $d$ represents embedding dimensionality, $m$ represents question length, and $n$ represents context length. To this end, the first step is to calculate the similarity between the contextual embedding of each context word, $u_{c_i}$ (i.e. the $i$-th column of $U_C$), and that of each question word, $u_{q_j}$ (i.e. the $j$-th column of $U_Q$). A similarity function for this purpose, which was originally proposed in BiDAF \cite{seo2016bidirectional} and has been adopted in many other MRC models \cite{yu2018qanet,pan2017memen,clark2017simple,wang2018multi,nishida2019multi,nishida2019answering}, is as follows:
\[s(u_{c_i}, u_{q_j}) = w_s^\top \ [u_{c_i}; u_{q_j}; u_{c_i} \odot u_{q_j}] \in \mathbb{R}\]
where $w_s$ is a trainable weight vector, $\odot$ represents element-wise multiplication, and $[;]$ represents vector concatenation. Besides, another similarity function, which has also been widely adopted \cite{hu2017reinforced,liu2017stochastic,wangwei2018multi,yan2019deep}, is as follows:
\[s(u_{c_i}, u_{q_j}) = \mathrm{ReLU}(W_s \ u_{c_i})^\top \ \mathrm{ReLU}(W_s \ u_{q_j}) \in \mathbb{R}\]
where $\mathrm{ReLU}()$ represents performing rectified linear activation, and $W_s$ is a trainable weight matrix. By applying any similarity function described above to every $(u_{c_i}, u_{q_j})$ pair, a similarity matrix $S \in \mathbb{R}^{n \times m}$ can be obtained, where each element $S_{i,j} = s(u_{c_i}, u_{q_j})$. On this basis, the next step is to calculate the context-attended contextual embedding summaries of the question, $\tilde{U}_{Q}$, and the question-attended contextual embedding summaries of the context, $\tilde{U}_{C}$, which can be obtained by applying the following equations separately \cite{seo2016bidirectional,pan2017memen,clark2017simple}:
\[\tilde{U}_{Q} = U_Q \ \mathrm{Softmax}_r^\top(S) \in \mathbb{R}^{d \times n}\]
\[\tilde{U}_{C} = \mathrm{Tile}_r(U_C \ \mathrm{Softmax}(\mathrm{Max}_r(S)), n) \in \mathbb{R}^{d \times n}\]
where $\mathrm{Softmax}_r()$ represents performing softmax normalization along the row dimension, $\mathrm{Tile}_r(x,y)$ represents tiling a vector $x$ for $y$ times along the row dimension, and $\mathrm{Max}_r()$ represents taking the maximum element of each row. Besides, $\tilde{U}_{C}$ can also be obtained by applying the following equation \cite{xiong2016dynamic,yu2018qanet}:
\[\tilde{U}_{C} = U_C \ \mathrm{Softmax}_c(S) \ \mathrm{Softmax}_r^\top(S) \in \mathbb{R}^{d \times n}\]
where $\mathrm{Softmax}_c()$ represents performing softmax normalization along the column dimension. With $\tilde{U}_{Q}$ and $\tilde{U}_{C}$ obtained as described above, the final step is to fuse them with $U_C$, which can be achieved by concatenating the matrices $U_C$, $\tilde{U}_{Q}$, $U_C \odot \tilde{U}_{Q}$, and $\tilde{U}_{C} \odot \tilde{U}_{Q}$ together along the column dimension \cite{seo2016bidirectional,yu2018qanet,clark2017simple}. Alternatively, the matrix concatenation can also be performed over $U_C$, $\tilde{U}_{Q}$, and $\tilde{U}_{C}$ \cite{xiong2016dynamic}, or simply over $U_C$ and $\tilde{U}_{Q}$ when $\tilde{U}_{C}$ is not available \cite{liu2017stochastic}. \\
What is described above is just a general mutual-matching attention mechanism, which is more or less different from those for practical MRC models. For example, SLQA+ \cite{wangwei2018multi} applies the following equation to obtain $\tilde{U}_{C}$:
\[\tilde{U}_{C} = U_C \ \mathrm{Softmax}_c(S) \in \mathbb{R}^{d \times n}\]
and also fuses $\tilde{U}_{C}$ with $U_Q$ as well as $\tilde{U}_{Q}$ with $U_C$ by applying a gated fusion method. Besides, Match-LSTM \cite{wang2016machine} uses an attention-wrapped RNN to process $U_C$ based on $U_Q$. Specifically, given the input $u_{c_i}$ and the hidden state $h_{i-1}$ at each time step, it first calculates $z_i$, an attended summary of $U_Q$, as follows:
\[t_j = w_z^\top \ \mathrm{Tanh}(W_Q u_{q_j} + W_C u_{c_i} + W_H h_{i-1}) \in \mathbb{R}\]
\[z_i = U_Q \ \mathrm{Softmax}(\{t_1, \ldots, t_m\}) \in \mathbb{R}^{d}\]
where $w_z$ is a trainable weight vector, $\mathrm{Tanh}()$ represents performing hyperbolic tangent activation, and $W_Q$, $W_C$, and $W_H$ are trainable weight matrices. Then it updates the hidden state with both $u_{c_i}$ and $z_i$ as follows:
\[h_t = \mathrm{RNN}(h_{i-1}, [u_{c_i};z_i])\]
Based on Match-LSTM, R-NET \cite{wang2017gated} modifies the attention-wrapped RNN by adding a gate to the update of the hidden state as follows:
\[h_t = \mathrm{RNN}(h_{i-1}, \mathrm{Sigmoid}(W_G [u_{c_i};z_i]) \odot [u_{c_i};z_i])\]
where $\mathrm{Sigmoid}()$ represents performing sigmoid activation, and $W_G$ is a trainable weight matrix. As a result, the update of the hidden state is somewhat affected by the relevance of the current context to the question.

\subsection{Self-matching Attention}
As the substantial component of the self-matching layer, self-matching attention is aimed at fusing the coarse memories of the context, $V_C \in \mathbb{R}^{d \times n}$, into themselves, where $d$ represents memory dimensionality, and $n$ represents context length. Self-matching attention was originally proposed in R-NET \cite{wang2017gated}, which uses a gated attention-wrapped RNN to process $V_C$ based on themselves following the same approach as in its mutual-matching attention. However, some other MRC models implemented self-matching attention following a similarity-summary-fusion approach \cite{clark2017simple,liu2017stochastic,wangwei2018multi}. Specifically, they first apply a similarity function, which is the same as or similar to those applied to their mutual-matching attention, to obtain a similarity matrix $D \in \mathbb{R}^{n \times n}$, where each element $D_{i,j}$ is the similarity between the coarse memory of the $i$-th context word, $v_{c_i}$ (i.e. the $i$-th column of $V_C$), and that of the $j$-th context word, $v_{c_j}$ (i.e. the $j$-th column of $V_C$). Then they calculate the self-attended coarse memory summaries of the context as $\tilde{V}_{C} = V_C \ \mathrm{Softmax}_c(D) \in \mathbb{R}^{d \times n}$. Here the diagonal elements of $D$ are usually set to $-\infty$ so that each context word will be forced to attend to the other context words rather than itself. Finally they fuse $\tilde{V}_{C}$ with $V_C$ by applying a fusion method, which is the same as or similar to those applied to their mutual-matching attention.

\subsection{Multi-round Attention}
Multi-round attention refers to performing both mutual-matching attention and self-matching attention repeatedly, which is aimed at effectively capturing the complicated interactions among the words in each given context-question pair. Some MRC models, such as FusionNet \cite{huang2017fusionnet} and DCN+ \cite{xiong2017dcn+}, implement multi-round attention by using the word representations output by the previous attention round (i.e. mutual-matching attention followed by self-matching attention) as the inputs to the current attention round. However, since different attention rounds share attentive information only through word representations, it is easy to cause attention redundancy and attention deficiency. For this reason, R.M-Reader \cite{hu2017reinforced} further uses the similarity matrices obtained in the previous attention round to refine the similarity matrices in the current attention round.

\section{Performance-boosting Approaches for MRC Models}
This section introduces some representative performance-boosting approaches for MRC models.

\subsection{Linguistic Embeddings}
It is both simple and effective to extend the lexical embeddings of the input layer with the corresponding linguistic embeddings. For example, DrQA \cite{chen2017reading} and SAN \cite{liu2017stochastic} extend their lexical embeddings with POS embeddings and NER embeddings. Besides, SEST \cite{liu2017structural} extends its lexical embeddings with structural embeddings based on constituency trees and dependency trees.

\subsection{Multi-round Reasoning}
Given a difficult document, human beings can finally understand it by reading it over and over again. This strategy can also be implemented to better address difficult MRC tasks, which is known as multi-round reasoning. For example, ReasoNet \cite{shen2017reasonet} uses reinforcement learning to dynamically determine the number of reasoning rounds. Besides, SAN \cite{liu2017stochastic} fixes the number of reasoning rounds but uses stochastic dropout in the output layer to avoid step bias.

\subsection{Reinforcement Learning}
To reward the answers that are textually similar to but positionally different from the golden answer, a reinforcement learning loss, which is measured according to the overlap between the predicted answer and the golden answer, can be combined with the traditional cross entropy loss to form a joint loss to optimize. This strategy was originally proposed in DCN+ \cite{xiong2017dcn+} and has been adopted in R.M-Reader \cite{hu2017reinforced}. Besides, SLQA+ \cite{wangwei2018multi} optimizes the traditional cross entropy loss in its pre-training stage and the joint loss in its fine-tuning stage.

\subsection{Data Augmentation}
Augmenting training examples can improve the performance of MRC models. For example, GDANs \cite{yang2017semi} uses a separate generative model to generate questions based on unlabeled text, which substantially improves its performance. Besides, QANet \cite{yu2018qanet} uses a separate back-and-forth translation model to paraphrase training examples, which brings its performance a significant gain.

\chapter{Transfer Learning in MRC Models}
From the perspective of MRC, transfer learning is aimed at incorporating text-style knowledge contained in external corpora (e.g. Wikipedia) into neural networks of MRC models. In practice, such transfer learning falls into two categories, namely feature-based transfer learning and fine-tuning-based transfer learning. They both require pre-training a source model through a foundation-level NLP task based on external corpora, but they are different in the way the pre-trained source model is merged into the target MRC model. In feature-based transfer learning, the pre-trained source model is only used as a supplement to the input layer of the target MRC model to generate additional features, which does not affect the design for the downstream layers. In fine-tuning-based transfer learning, on the contrary, the pre-trained source model is directly used as the target MRC model just with a slight change made to the output layer, therefore training the target MRC model can be considered as fine-turning the pre-trained source model. This chapter introduces some representative applications of both feature-based transfer learning and fine-tuning-based transfer learning in MRC models.

\section{Feature-based Transfer Learning}
In feature-based transfer learning, the source model takes a sequence of words as inputs and generates the corresponding word representations before outputting for the foundation-level NLP task. After pre-trained based on the external corpora, the source model can be used in the input layer of the target MRC model to generate word representations for each given context-question pair. The resulting word representations are known as contextualized word representations, since the representation of each word is not constant but dynamically determined according to the contextual information. This section introduces two representative applications of feature-based transfer learning, namely Context Vectors (CoVe) \cite{mccann2017learned} and Embeddings from Language Models (ELMo) \cite{peters2018deep}.

\subsection{Context Vectors}
In Context Vectors (CoVe) \cite{mccann2017learned}, the source model is a sequence-to-sequence machine translation model, which consists of a two-layer BiLSTM encoder and a two-layer attention-wrapped LSTM decoder. For the pre-training of the source model, the foundation-level NLP task is to translate an English sentence into German, and the external corpora are from the corresponding WMT datasets. In the target MRC model, the pre-trained source model drops the decoder and takes the outputs of the encoder as the contextualized word representations, which are concatenated with the original word representations of the input layer. Some MRC models, such as DCN \cite{xiong2016dynamic} and SAN \cite{liu2017stochastic}, have benefit from CoVe. However, since the pre-training of the source model is a supervised learning process, where the external corpora are expensive, the effectiveness of CoVe is limited.

\subsection{Embeddings from Language Models}
In Embeddings from Language Models (ELMo) \cite{peters2018deep}, the source model is a three-layer bidirectional language model, where the bottom layer is a character-level CNN and the upper two layers are BiLSTMs. For the pre-training of the source model, the foundation-level NLP task is to predict the next word and the previous word at each time step based on the output of the top-layer forward LSTM and that of the top-layer backward LSTM separately, and the external corpora are from 1 Billion Word Language Model Benchmark. In the target MRC model, a position-wise weighted sum of the outputs of the three layers of the pre-trained source model is first calculated as the contextualized word representations, which are then scaled and concatenated with the original word representations of the input layer. Many MRC models have benefit from ELMo, and it has been proved that in some RNN-based MRC models, such as BiDAF\cite{seo2016bidirectional} and DocumentQA \cite{clark2017simple}, applying the above operations also to the feature extraction layer can further improve the performance. Due to the fact that the pre-training of the source model is an unsupervised learning process, it is convenient and economical to collect the external corpora. However, since LSTM has a bottle neck in capturing long-term dependencies from sequential data, the effectiveness of ELMo is still limited.

\section{Fine-tuning-based Transfer Learning}
In fine-tuning-based transfer learning, the source model adopts a well-designed input format, which is compatible with various NLP tasks, such as MRC. Besides, by leveraging Transformer \cite{vaswani2017attention}, which is excellent at capturing long-term dependencies from sequential data, the source model can generate profound word representations before outputting for the foundation-level NLP task. As a result, after pre-trained based on the external corpora and slightly changed in the output layer, the source model can be directly used as the target MRC model, a fine-tuning of which based on the target MRC dataset can lead to desirable performance. This section introduces two representative applications of fine-tuning-based transfer learning, namely Generative Pre-training for Transformers (GPT) \cite{radford2018improving} and Bidirectional Encoder Representations from Transformers (BERT) \cite{devlin2018bert}.

\subsection{Generative Pre-training for Transformers}
In Generative Pre-training for Transformers (GPT) \cite{radford2018improving}, the source model is a unidirectional language model, which consists of multiple Transformer decoder layers. For the pre-training of the source model, the foundation-level NLP task is to predict the next word based on the output of the top-layer Transformer decoder at each position, and the external corpora are from BooksCorpus and 1 Billion Word Language Model Benchmark. For the training of the target MRC model, which is actually the fine-tuning of the pre-trained source model, each example in the target MRC dataset is converted to an input sequence that starts with a $\langle s \rangle$ token and ends with a $\langle e \rangle$ token, with the context and the question being the sub-sequences separated by a $\$$ token. Due to the strong ability of Transformer in capturing long-term dependencies from sequential data and the large scale of the external corpora, the pre-trained source model have obtained enough external knowledge, which enables it to achieve very good performance on the target MRC task through a simple fine-tuning. However, since the source model only allows each word to depend on its previous words, which causes the dependencies in the opposite direction to be neglected, the effectiveness of GPT is limited.

\subsection{Bidirectional Encoder Representations from Transformers}
In Bidirectional Encoder Representations from Transformers (BERT) \cite{devlin2018bert}, the source model is a bidirectional language model, which consists of multiple Transformer encoder layers. Each input sequence taken by the source model starts with a $[CLS]$ token, which is followed by two sub-sequences, with each one ending with a $[SEP]$ token. For the pre-training of the source model, the foundation-level NLP task is composed of two sub-tasks, the first one is to predict some randomly masked words in the input sequence based on the outputs of the top-layer Transformer encoder at the unmasked positions, and the second one is to determine whether the two sub-sequences are consecutive based on the output of the top-layer Transformer encoder at the position of the $[CLS]$ token. The external corpora used for the pre-training are from BooksCorpus and Wikipedia. For the training of the target MRC model, which is actually the fine-tuning of the pre-trained source model, each example in the target MRC dataset is converted to an input sequence by using the question as the first sub-sequence and the context as the second sub-sequence. BERT-based MRC models have outperformed human beings in several MRC tasks, this is not only due to the strong ability of Transformer in capturing long-term dependencies from sequential data and the large scale of the external corpora, but also due to the smart design of the foundation-level NLP task, where the first sub-task allows the dependencies in both directions to be utilized, and the second sub-task further allows the inter-sentence dependencies to be utilized. However, since BERT-based MRC models are very large in scale, their pre-training, fine-tuning, and inference are computationally expensive for most users.

\chapter{Knowledge Base Encoding in MRC Models}
Knowledge bases, which originate from expert systems, are aimed at storing knowledge in structured forms. Usually, a knowledge base physically or logically organizes its knowledge as a set of ``subject-predicate-object'' triples, where each triple refers to a fact that the subject is related to the object through the predicate. As a result, by representing the subjects and objects as vertices and the predicates as edges, the knowledge contained in a knowledge base can be expressed as a graph. Nowadays, a broad variety of knowledge bases with graph-style knowledge are available, such as WordNet \cite{miller1995wordnet} with semantic knowledge, ConceptNet \cite{speer2017conceptnet} with commonsense knowledge, and Freebase \cite{bollacker2008freebase} with factoid knowledge. \\
By reasoning over the graph-style knowledge, a knowledge base can be directly used to answer certain real-world questions (e.g. ``which city is the capital of Canada''), which is known as knowledge base question answering (KBQA). However, unlike in KBQA, the question answering in MRC has to be based on a given context rather than external knowledge bases. Even so, since background knowledge always plays a positive role in the reading comprehension of human beings, it will be beneficial if the graph-style knowledge contained in external knowledge bases is incorporated as background knowledge into the neural networks of MRC models. An effective way to achieve this goal is to encode external knowledge bases in vector space such that the encoding outputs can be used to enhance word representations. A word embedding refinement approach \cite{weissenborn2017dynamic} first retrieves from ConceptNet the triples relevant to each given context-question pair and transforms them into text-form assertions, then passes the assertions through a BiLSTM-based encoding layer, and finally uses the encoding outputs to update the corresponding word embeddings. A knowledge memory approach \cite{mihaylov2018knowledgeable} is somewhat similar, but it encodes each obtained assertion into a key-value memory, calculates an attended memory summary for each word with the word representation as the query, and combines the resulting summary to the word representation. An entity description encoding approach \cite{long2017world} first extracts from Freebase the text-form entity descriptions of the entities existing in each given context-question pair, then passes the entity descriptions through an LSTM-based encoding layer, and finally uses the last encoding output for each entity description as the representation of the corresponding entity. \\
Compared with the transfer learning approaches introduced in the previous chapter, which have significantly promoted the utilization of text-style knowledge contained in external corpora, the knowledge base encoding approaches introduced here have not brought any breakthrough to the utilization of graph-style knowledge contained in external knowledge bases. The reason for this is two-fold, on the one hand, only certain isolated elements (i.e. triples and entity descriptions) of the external knowledge bases are involved, but the structural features of the graph-style knowledge are ignored; on the other hand, the encoding is not based on the graph-style knowledge itself, but certain forms of text derived from it.

\chapter{Conclusion}
This report is a survey on the existing tasks and models in MRC. For MRC tasks, some representative simple-reasoning and complex-reasoning tasks are introduced, where the emphasis is on dataset collection and performance evaluation. For MRC models, the primary aspects of model development are introduced, which include architecture designs, attention mechanisms, and performance-boosting approaches. On this basis, some recently proposed approaches to utilizing external knowledge in MRC models are introduced, which include transfer learning approaches to utilizing text-style knowledge contained in external corpora and knowledge base encoding approaches to utilizing graph-style knowledge contained in external knowledge bases. Since the knowledge base encoding approaches are often barely satisfactory in practice, more efforts are still required to improve the utilization of graph-style external knowledge. For this purpose, the following open problems will be carefully considered in the future research:
\begin{itemize}
\item The knowledge base encoding approaches are an implicit way to utilize graph-style external knowledge, since with these approaches human beings can neither understand nor control the functioning of graph-style external knowledge in MRC models. Can graph-style external knowledge be utilized in an explicit (i.e. understandable and controllable) way?
\item Instead of encoding the text derived from isolated elements of external knowledge bases, is there a way to convert each given context-question pair, which is originally in a text form, into a graph of words, sentences, or even passages, by leveraging graph-style external knowledge? If so, how should MRC models use the resulting graph to get the answer?
\item Are there any specific situations where properly utilizing graph-style external knowledge can supplement or outperform the transfer learning approaches to utilizing text-style external knowledge?
\end{itemize}

\bibliographystyle{report}
\bibliography{report}
\end{document}